\documentclass{article}

\usepackage{arxiv}

\usepackage[utf8]{inputenc} 
\usepackage[T1]{fontenc}    
\usepackage{url}            
\usepackage{booktabs}     
\usepackage{amsfonts}       
\usepackage{nicefrac}       
\usepackage{microtype}      %
\usepackage{lipsum}		
\usepackage{amsmath}
\usepackage{amssymb}
\usepackage{amsthm}
\usepackage{graphicx}
\usepackage{float}
\usepackage[normalem]{ulem}
\useunder{\uline}{\ul}{}
\usepackage{subcaption}
\usepackage[numbers]{natbib}
\usepackage{listings}
\usepackage{thmtools}
\usepackage{hyperref}

\usepackage{tikz}
\usepackage{titlesec}
\usepackage{fullpage}
\usepackage{lastpage}
\usepackage{algorithm} 
\usepackage[noend]{algpseudocode} 

\numberwithin{figure}{section}
\title{Exponential Tilting of Generative Models: Improving Sample Quality by Training and Sampling from Latent Energy
}

\author{
  Zhisheng Xiao \thanks{equal contribution} \\
  Computational and Applied Mathematics\\
  University of Chicago\\
  Chicago, IL, 60637\\
  \texttt{zxiao@uchicago.edu} \\
   \And
  Qing Yan \footnotemark[1]\\
  Department of Statistics\\
  University of Chicago\\
  Chicago, IL, 60637\\
  \texttt{yanq@uchicago.edu} \\
    \And
  Yali Amit \\
  Department of Statistics\\
  University of Chicago\\
  Chicago, IL, 60637\\
  \texttt{amit@marx.uchicago.edu} \\
}

\begin{document}
\maketitle

\begin{abstract}
In this paper, we present a general method that can improve the sample quality of pre-trained likelihood based generative models. Our method constructs an energy function on the latent variable space that yields an energy function on samples produced by the pre-trained generative model. The energy based model is efficiently trained by maximizing the data likelihood, and after training, new samples in the latent space are generated from the energy based model and passed through the generator to producing samples in observation space. We show that using our proposed method, we can greatly improve the sample quality of popular likelihood based generative models, such as normalizing flows and VAEs, with very little computational overhead.  
\end{abstract}

\section{Introduction}\label{intro}
Recent advances in deep likelihood based generative models \cite{vae,rezende, pixelcnn, pixelcnn++, glow} enable the modeling of very high dimensional and complicated data such as natural images, sequences \cite{wavenet} and graphs \cite{graphvae}. Compared to Generative Adversarial Networks (GANs) \cite{goodfellow2014generative}, these models can evaluate the likelihood of input data easily, return latent variables that are useful for downstream tasks and do not suffer from the instability and mode dropping issues \cite{salimans2016improved} of GAN training. However, GANs are still the state-of-the-art generative models for many generative tasks, because they can produce sharper and more realistic samples than non-adversarial models \cite{brock2018large}. Much effort has been devoted to improving the sample quality of non-adversarial, likelihood based generative models \cite{van2017neural,behrmann2018invertible,huang2020augmented,dai2019diagnosing} by modifying the training objectives. 

This paper aims at a slightly different task: we want to improve the sample quality of an existing generative model through a better sampling procedure. Note that in deep generative models, samples are usually obtained by sending latent variables through a deterministic transformation, where the latent variables are sampled from a pre-defined prior distribution. Controlling the temperature of sampling from the prior may produce better samples, but this comes at the cost of less diversity \cite{glow}. Recently, there is a line of literature that improves the sample quality of pre-trained GANs \cite{tanaka2019discriminator,che2020your}. They utilize the information contained in the discriminator of GANs to obtain better samples of the latent variable. In particular, \cite{che2020your} sample latent variables from an energy-based model (EBM) defined jointly by the generator and discriminator. 

Extension of these ideas to likelihood based models is nontrivial, because the discriminator of GANs is critical to provide guidance for moving the latent variables. We extend these methods to generative models without adversarial training by constructing a latent variable EBM that consists of the pre-trained generative model and an energy function. The EBM can be trained efficiently by maximizing the data likelihood, and we observe that training the EBM only adds a slight computational overhead, as the convergence is fast. After convergence, new samples are produced by latent variables sampled from the EBM. We show that our method can effectively improve the sample quality of a variety of pre-trained generative models, including normalizing flows, VAEs and a combination of the two. 

\section{Background on Energy-Based Models}
An Energy-based Model assumes a gibbs distribution 
\begin{align}
\label{EBM}
    p_{\theta}(\mathbf{x})=\frac{\exp \left(-E_{\theta}(\mathbf{x})\right)}{Z_{\theta}}
\end{align}
over data $\mathbf{x} \in \mathcal{X}$. $E_{\theta}(\mathbf{x})$ is the energy function with parameter $\theta$, and $Z_{\theta}=\int_{\mathbf{x}} \exp \left(-E_{\theta}(\mathbf{x})\right)$ is the normalizing constant. When $\mathbf{x}$ is an image, $E_{\theta}(\mathbf{x})$ is usually chosen to be a convolutional neural network with scalar output \cite{du2019implicit}. The EBM can be trained by the maximum likelihood principle, namely minimizing the negative log likelihood $L(\theta) = \mathbb{E}_{\mathbf{x} \sim p_{D}}\left[-\log p_{\theta}(\mathbf{x})\right]$. This objective is known to have derivative 
\begin{align}
\label{derivative}
    \frac{\partial L(\theta)}{\partial \theta} = \mathbb{E}_{p_{D}\left(\mathbf{x}\right)}\left[\frac{\partial E_{\theta}\left(\mathbf{x}\right)}{\partial \theta}\right]-\mathbb{E}_{p_{\theta}\left(\mathbf{x}^{\prime}\right)}\left[\frac{\partial E_{\theta}\left(\mathbf{x}^{\prime}\right)}{\partial \theta}\right]
\end{align}
where $\mathbf{x}^{\prime}$ represents a sample drawn from the EBM. However, it is difficult to draw samples from such complex unnormalized distributions, and one typically needs to employ MCMC algorithms. One efficient MCMC
algorithm in high dimensional continuous state spaces is Stochastic Gradient Langevin dynamics popularized in the statistics literature in the early 90's in \cite{amit-jasa} and introduced in the deep learning literature in \cite{10.5555/3104482.3104568}. This algorithm initializes at $\mathbf{x}_0$ and runs updates
\begin{align}
\label{LD}
    \mathbf{x}_{i+1}=\mathbf{x}_{i}-\frac{\epsilon}{2} \nabla_{\mathbf{x}} E_\theta(\mathbf{x})+\sqrt{\epsilon} \mathbf{\omega}, \mathbf{\omega} \sim \mathcal{N}(0, \mathbf{I}).
\end{align}
The continuous Langevin dynamics is guaranteed to produce samples from the target distribution. In practice we use a discrete approximation, which yields a Markov chain with invariant distribution close to the original target distribution. 

\section{Methodology}
\subsection{Exponential tilting of Generative Models}
Suppose we have a pre-trained probabilistic generative model $p_{\phi^*}(\mathbf{x})$ over data space $\mathcal{X}$, we can define a new model by ``exponential tilting" with energy function $E_{\theta}(\mathbf{x})$: 
\begin{equation}
\label{ET}
    p_{\phi^*,\theta}(\mathbf x) = \frac{p_{\phi^*}(\mathbf x)\exp{(-E_\theta(\mathbf x))}}{Z_{\phi^*,\theta}},
\end{equation}
where $Z_{\phi^*,\theta} = \int p_{\phi^*}(\mathbf x)\exp{(-E_\theta(\mathbf x))} d\mathbf x$ is the corresponding normalizing constant. Since the plain EBM (\ref{EBM}) is a special case with $p_{\phi^*} = \text{const}$, we can apply the training strategies in (\ref{derivative}), (\ref{LD}) to $p_{\phi^*,\theta}(\mathbf{x})$.
The objective $L(\theta) = \mathbb{E}_{\mathbf{x} \sim p_{D}}\left[-\log p_{\phi^*,\theta}(\mathbf{x})\right]$ has gradient
\begin{align}
\label{derivative::ET}
    \frac{\partial L(\theta)}{\partial \theta} = \mathbb{E}_{p_{D}\left(\mathbf{x}\right)}\left[\frac{\partial E_{\theta}\left(\mathbf{x}\right)}{\partial \theta}\right]-\mathbb{E}_{p_{\phi^*,\theta}\left(\mathbf{x}^{\prime}\right)}\left[\frac{\partial E_{\theta}\left(\mathbf{x}^{\prime}\right)}{\partial \theta}\right],
\end{align}
and the Langevin dynamics to generate samples from $p_{\phi^*,\theta}(\mathbf{\mathbf{x}})$ is 
\begin{align*}
    \mathbf{x}_{i+1} = \mathbf{x}_{i}-\frac{\epsilon}{2} \nabla_{\mathbf{x}} \left(E_\theta(\mathbf{x})-\log p_{\phi^*}(\mathbf{x})\right)+\sqrt{\epsilon} \mathbf{\omega}.
\end{align*}

Note that the derivative of the log likelihood of the original generative model $p_{\phi^*}(\mathbf{x})$ appears in the update, driving the Langevin dynamics to samples $\mathbf{x}$ with large likelihood and low energy simultaneously. However, this only works for generative models with tractable likelihood. More importantly, operating in the pixel space may be inefficient as it completely ignores the latent variables.

\subsection{EBM in Latent Space}\label{latent ebm}
Many types of probabilistic generative models, including normalizing flows and VAEs, adopt a decoder structure in their generation process, namely there is a pre-defined prior distribution $p(\mathbf{z})$, and samples are generated by
\begin{align*}
    \mathbf{z} \sim p(\mathbf{z}), \quad \mathbf{x} = G_{\phi^*}(\mathbf{z}).
\end{align*}
We can therefore re-parametrize the EBM $p_{\phi^*,\theta}(\mathbf{\mathbf{x}})$ in \eqref{ET} with the latent variable $\mathbf{z}$ and obtain
\begin{equation}\label{new EBM}
    p_{\phi^*,\theta}(\mathbf{z}) = \frac{p(\mathbf{z})\exp{(-E_\theta(G_{\phi^*}(\mathbf{z})))}}{Z_{\phi^*,\theta}}.
\end{equation}
When $p_{\phi^*,\theta}(\mathbf{z})$ is trained by maximizing the likelihood, the second term of \eqref{derivative} can also be re-parametrized to z space:
\begin{align}
\label{reparam}
    \mathbb{E}_{p_{\phi^*,\theta}\left(\mathbf{x}\right)}\left[\frac{\partial E_{\theta}\left(\mathbf{x}\right)}{\partial \theta}\right] =  \mathbb{E}_{p_{\phi^*,\theta}(z)}\left[\frac{\partial E_\theta(G_{\phi^*}(z))}{\partial \theta}\right]
\end{align}
See Appendix \ref{proof1} for a simple derivation.  Similarly, samples from $p_{\phi^*,\theta}(\mathbf{z})$ can be obtained through running the Langevin dynamics
\begin{align}\label{z ld}
    \mathbf{z}_{i+1} = \mathbf{z}_{i}-\frac{\epsilon}{2} \nabla_{\mathbf{z}} \left(E_\theta(G_{\phi^*}(\mathbf{z}))-\log p(\mathbf{z})\right)+\sqrt{\epsilon} \mathbf{\omega}.
\end{align}
Sometimes $\mathbf{z}$ has lower dimensionality than $\mathbf{x}$ and $p(\mathbf{z})$ can often be computed easily, therefore training and sampling from $p_{\phi^*,\theta}(\mathbf{z})$ is more efficient.


\section{Related Work}\label{rw}
Our work is closely related to recent literature that uses a discriminator to improve the sample quality of an existing GAN. Discriminator rejection
sampling \cite{azadi2018discriminator} and Metropolis-Hastings
GANs \cite{turner2018metropolis} use the discriminator as a criterion of accepting or rejecting samples from the generator. They are inefficient as many of samples may be rejected. Discriminator optimal transport (DOT) \cite{tanaka2019discriminator} and Discriminator Driven Latent Sampling (DDLS) \cite{che2020your} both move the latent variable to make samples better fool the discriminator. In particular, \cite{tanaka2019discriminator} uses deterministic gradient descent in the latent space, while \cite{che2020your} formulates an EBM on latent variables and use Langevin dynamics to sample latent variables. All these methods rely on the fact that the discriminator of a trained GAN is a good classifier on real/fake images, which is not applicable to likelihood based generative models.

When applied on VAEs, our method shares similarity with some recent literature that trains an auxiliary model to match the empirical latent distribution of an existing VAE, and samples are produced by latent variables generated by the auxiliary model. The auxiliary model can be another VAE \cite{dai2019diagnosing}, normalizing flow \cite{xiao2019generative} or auto-regressive model \cite{van2017neural}. Our method use an EBM as the auxiliary model, but its purpose is to define the energy for generated samples. More importantly, other methods can improve the sample quality only when increasing the weight on the reconstruction term in the objective, which will make the latent representation less structured. In contrast, our method can improve the sample quality of VAE trained without modifying the objective.

Our method heavily relies on the progress of training deep EBMs. Recently, \cite{du2019implicit,nijkamp2019anatomy,nijkamp2019learning} successfully scales up the maximum likelihood learning of EBMs to high dimensional images. In particular, we follow \cite{nijkamp2019anatomy,nijkamp2019learning} and use short run non convergent MCMC to train our EBMs. 

\section{Experiments}
\subsection{Toy dataset}
To give a quick proof-of-concept, we apply our method on toy datasets (25-Gaussians and Swiss Roll) following the setting of \cite{tanaka2019discriminator}. We first train a VAE on the training data, and then we fix the VAE and train a latent EBM as described in Section \ref{latent ebm}. The decoder and the energy function (which corresponds to the discriminator in GANs) have simple fully connected structure as described in \cite{tanaka2019discriminator}. Note that we do not use normalizing flows on toy datasets, because vanilla flow is heavily constrained by the manifold structure
of the prior distribution, making it very hard to model distributions like the 25-Gaussians.

\begin{figure}[ht]
    \centering
    \begin{subfigure}{.95\linewidth}
    \includegraphics[scale=0.6]{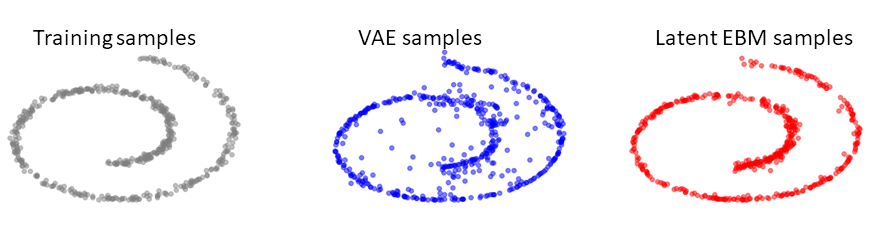}
        \caption{Swiss Roll}
    \end{subfigure}
    \vskip1em
    \begin{subfigure}{.95\linewidth}
    \includegraphics[scale=0.6]{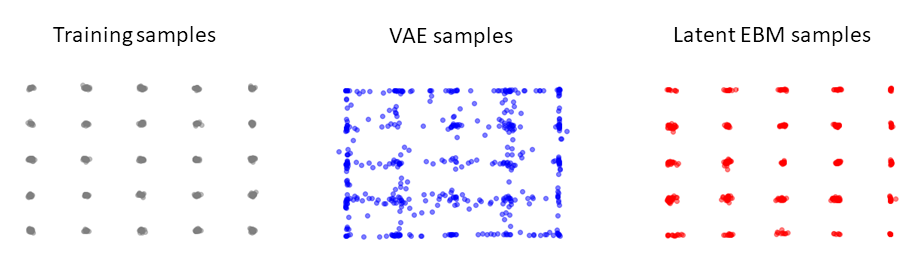}
       \caption{25-Gaussians}
    \end{subfigure}
    \caption{\label{toy}
     Applying latent EBM to VAEs trained on Swiss Roll and 25-Gaussians datset.}
\end{figure}

We show qualitative results in Figure \ref{toy}. We observe that although samples from VAEs can basically cover the shape of the true distribution, many samples still appear at low density regions. In contrast, by sampling and decoding latent variables obtained from the post-trained latent EBM, we can accurately preserve all modes in the target distribution while eliminating spurious modes in the 25-Gaussians case. In the Swiss Roll case, it is also clear that the EBM better captures the underlying data distribution. 

\subsection{Image dataset}
In this section we evaluate the performance of the proposed latent EBM on MNIST, Fashion MNIST and CIFAR-10 dataset. We use different decoder based generative models $G_{\phi^*}(\mathbf{z})$, including normalizing flow, VAE and GLF, which uses a latent flow model \cite{xiao2019generative} that combines a deterministic auto-encoder and a normalizing flow on the latent variables. Note that our main focus is on the relative improvements of sampling from the EBMs over sampling from base generative models, and therefore the performances of the base generative models may not be state-of-the-art. In fact, we adopt relatively simple network structures for convenience. 

As in \cite{du2019implicit}, we use a convolutional network with scalar outputs as $E_{\theta}$. We adopt short-run non-persistent MCMC, so the Langevin dynamics \eqref{z ld} is run with $\mathbf{z}_0$ initialized from $p(\mathbf{z})$ for a small number of steps. We fix $G_{\phi^*}(\mathbf{z})$ and $E_{\theta}$ is trained by maximum likelihood. For details on the settings of our experiments, see Appendix \ref{setting}.

\begin{figure}[h!]
    \centering
    \begin{subfigure}{.95\linewidth}
    \includegraphics[scale=1]{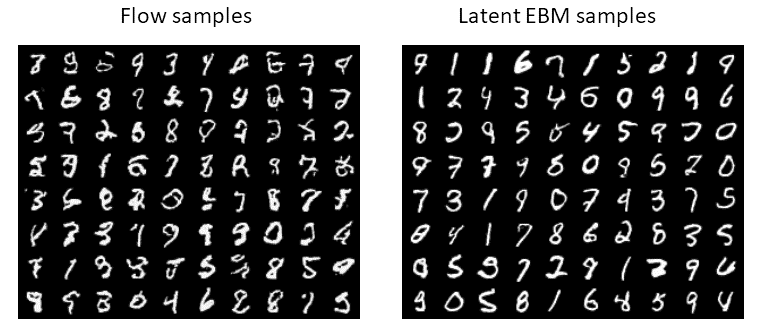}
    \end{subfigure}
    \begin{subfigure}{.95\linewidth}
    \includegraphics[scale=1]{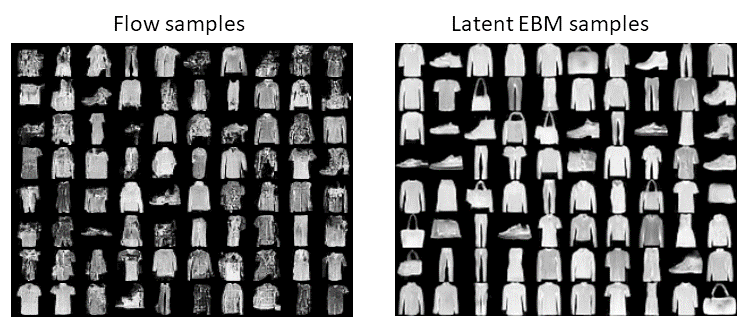}
    \end{subfigure}
    \begin{subfigure}{.99\linewidth}
    \includegraphics[scale=0.68]{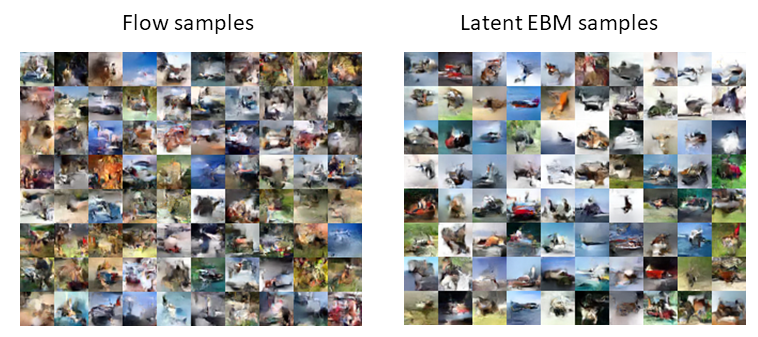}
    \end{subfigure}
    \caption{\label{glow fig}
     Applying latent EBM to GLOW trained on MNIST, Fashion and CIFAR-10. \textbf{Left}: samples generated by $\mathbf{z}$'s from the prior. \textbf{Right}: samples generated by $\mathbf{z}$'s from $ p_{\phi^*,\theta}(\mathbf{z})$. }
    
\end{figure}

\begin{figure}[ht]
    \centering
    \includegraphics[scale=0.6]{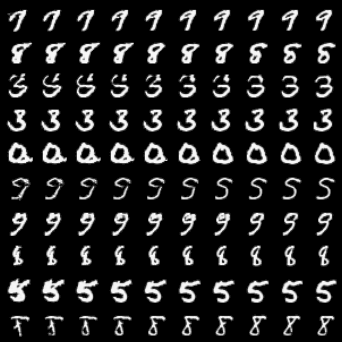}
    \caption{\label{traj fig}
     MNIST Langevin dynamics visualization, initialized at samples from prior (the leftmost column).}
\end{figure}

We show some qualitative results of training our proposed latent EBMs on top of a GLOW \cite{glow} model in Figure \ref{glow fig}. From Figure \ref{glow fig}, we clearly observe that samples generated by latent variables obtained from the latent EBMs have higher quality than samples from the base generative model (i.e., decoding latent variables from prior distribution). On MNIST and Fashion MNIST, samples obtained through the latent EBM have smoother shapes than samples from the GLOW. On CIFAR-10, the latent EBM effectively corrects the noisy backgrounds of the samples generated by the GLOW. We illustrate the process of Langevin dynamics sampling from the latent EBM in Figure \ref{traj fig}, where we generate samples for every 10 iterations. Apparently the the Langevin dynamics is going towards latent variables that produce more semantically meaningful and sharp samples.

More qualitative examples, including results of training latent EBMs on top of VAE and GLF are presented in Appendix \ref{qualitative}. In Figure \ref{vae large}, we observe that the VAE+latent EBM generates sharper samples than VAE alone. However, it should be noted that, since our EBM operates on the latent space, the overall sample quality is constrained by the capacity of the base generative models. 

Our observation on the improvements of sample quality can be confirmed by quantitative results in Table \ref{table:fid}, where we compare the FID scores \cite{heusel2017gans} of different models. We see that sampling latent variables from the latent EBM significantly improves the quality of generated samples over directly sampling from $p_{\theta}(\mathbf{x})$. In addition, we see that methods mentioned in Section \ref{rw} do not improve the sample quality of VAEs \textit{without posing a large weight on reconstruction term}, while our method can generate better samples without changing the VAE's objective, leading to good sample quality \textit{and} structured latent representation. For completeness, we also present results of training EBMs directly on pixel space using the same model structures. The results are in the same range as the latent EBM models, but we observe that training EBMs on data is more sensitive to hyper-parameter settings and more computationally expensive. 

\begin{table}
\centering  
\centering
   \begin{tabular}{llll}
   \toprule
   & MNIST  & Fashion & CIFAR-10 \\
   \midrule
   GLOW & 29.4  & 58.7 & 76.2 \\
   GLOW + EBM    & 12.3 & 41.6  & 67.8      \\
   \midrule
   VAE & 18.9 & 57.1 & 139.6     \\
   Two-stage VAE & 19.3  & 55.7  & 134.6    \\
   VAE + flow    & 18.6 & 52.3 & 128.2    \\
   VAE + EBM    & 16.0 & 38.1 & 108.4    \\
   \midrule
   GLF    & 14.2 & 32.5 & 96.6    \\
   GLF + EBM    & 12.1 & 25.3 & 85.1    \\
   \midrule
   EBM on data   &25.5  &39.3 &  80.6  \\

   \bottomrule
   \end{tabular}
\caption{Comparing the FID scores of base generative models and generative models + exponential tilting with latent EBMs. Scores are computed using 10000 generated samples and real samples from the test set.} \label{table:fid}
\end{table}

\subsection{Overfitting issue of training latent EBMs}
As pointed out in \cite{grathwohl2019your,nijkamp2019anatomy}, instability and overfitting are frequently observed when training Energy-based Models. Overfitting happens when the energy of training samples are much lower than samples drawn from the EBM, which causes the model to produce worse samples. We find heuristic approaches such as energy regularization and gradient clipping not helpful in preventing overfitting, so we simply stop the training of the latent EBM when sample quality deteriorates. We believe that future studies in improving the stability of EBM training can further boost the performance of our method.

\subsection{Training time}
One training step of EBM requires obtaining a sample by running multiple steps of MCMC, and therefore it is much slower than one training step of the base generative model. However, we find that very few training iterations are needed to achieve the results in Table \ref{table:fid}. Specifically, we only train latent EBMs for 200 steps when $G_{\phi^*}(\mathbf{z})$ is a GLOW or GLF, and 1000 steps when $G_{\phi^*}(\mathbf{z})$ is a VAE. Here a step refers to {\bf one} batch, not an entire epoch. These numbers are several orders of magnitude smaller than the training steps needed for the base generative models. Therefore, our method does not add much computational overhead. As a comparison, training EBMs on pixel space typically requires more than $100$k steps.

\section{Conclusion}
In this paper, we propose to train an Energy-based model on the latent space of pre-trained generative models. We show that with little computational overhead, we can improve the sample quality of a variety of generative models, including normalizing flow and VAE, by sampling latent variables from the EBM. Our method also provides a general framework that connects Energy-based models and other likelihood based generative models. We believe this connection is an interesting direction for future research. 
\newpage
\bibliographystyle{plain}
\bibliography{main}

\begin{thebibliography}{10}

\bibitem{amit-jasa}
Y.~Amit, U.~Grenander, and M.~Piccioni.
\newblock Structural image restoration through deformable template.
\newblock {\em Journal of the American Statistical Association},
  86(414):376--387, 1991.

\bibitem{azadi2018discriminator}
Samaneh Azadi, Catherine Olsson, Trevor Darrell, Ian Goodfellow, and Augustus
  Odena.
\newblock Discriminator rejection sampling, 2018.

\bibitem{behrmann2018invertible}
Jens Behrmann, Will Grathwohl, Ricky~TQ Chen, David Duvenaud, and
  J{\"o}rn-Henrik Jacobsen.
\newblock Invertible residual networks.
\newblock {\em arXiv preprint arXiv:1811.00995}, 2018.

\bibitem{brock2018large}
Andrew Brock, Jeff Donahue, and Karen Simonyan.
\newblock Large scale gan training for high fidelity natural image synthesis.
\newblock {\em arXiv preprint arXiv:1809.11096}, 2018.

\bibitem{che2020your}
Tong Che, Ruixiang Zhang, Jascha Sohl-Dickstein, Hugo Larochelle, Liam Paull,
  Yuan Cao, and Yoshua Bengio.
\newblock Your gan is secretly an energy-based model and you should use
  discriminator driven latent sampling.
\newblock {\em arXiv preprint arXiv:2003.06060}, 2020.

\bibitem{dai2019diagnosing}
Bin Dai and David Wipf.
\newblock Diagnosing and enhancing vae models.
\newblock {\em arXiv preprint arXiv:1903.05789}, 2019.

\bibitem{du2019implicit}
Yilun Du and Igor Mordatch.
\newblock Implicit generation and generalization in energy-based models.
\newblock {\em arXiv preprint arXiv:1903.08689}, 2019.

\bibitem{goodfellow2014generative}
IJ~Goodfellow, J~Pouget-Abadie, M~Mirza, B~Xu, D~Warde-Farley, S~Ozair,
  A~Courville, and Y~Bengio.
\newblock Generative adversarial networks. arxiv 2014.
\newblock {\em arXiv preprint arXiv:1406.2661}, 2014.

\bibitem{grathwohl2019your}
Will Grathwohl, Kuan-Chieh Wang, J{\"o}rn-Henrik Jacobsen, David Duvenaud,
  Mohammad Norouzi, and Kevin Swersky.
\newblock Your classifier is secretly an energy based model and you should
  treat it like one.
\newblock {\em arXiv preprint arXiv:1912.03263}, 2019.

\bibitem{heusel2017gans}
Martin Heusel, Hubert Ramsauer, Thomas Unterthiner, Bernhard Nessler, and Sepp
  Hochreiter.
\newblock Gans trained by a two time-scale update rule converge to a local nash
  equilibrium.
\newblock In {\em Advances in neural information processing systems}, pages
  6626--6637, 2017.

\bibitem{huang2020augmented}
Chin-Wei Huang, Laurent Dinh, and Aaron Courville.
\newblock Augmented normalizing flows: Bridging the gap between generative
  flows and latent variable models.
\newblock {\em arXiv preprint arXiv:2002.07101}, 2020.

\bibitem{vae}
Diederik~P Kingma and Max Welling.
\newblock Auto-encoding variational bayes.
\newblock {\em arXiv preprint arXiv:1312.6114}, 2013.

\bibitem{glow}
Durk~P Kingma and Prafulla Dhariwal.
\newblock Glow: Generative flow with invertible 1x1 convolutions.
\newblock In {\em Advances in Neural Information Processing Systems}, pages
  10215--10224, 2018.

\bibitem{graphvae}
Thomas~N Kipf and Max Welling.
\newblock Variational graph auto-encoders.
\newblock {\em arXiv preprint arXiv:1611.07308}, 2016.

\bibitem{nalisnick2018deep}
Eric Nalisnick, Akihiro Matsukawa, Yee~Whye Teh, Dilan Gorur, and Balaji
  Lakshminarayanan.
\newblock Do deep generative models know what they don't know?
\newblock {\em arXiv preprint arXiv:1810.09136}, 2018.

\bibitem{nijkamp2019anatomy}
Erik Nijkamp, Mitch Hill, Tian Han, Song-Chun Zhu, and Ying~Nian Wu.
\newblock On the anatomy of mcmc-based maximum likelihood learning of
  energy-based models.
\newblock {\em arXiv preprint arXiv:1903.12370}, 2019.

\bibitem{nijkamp2019learning}
Erik Nijkamp, Mitch Hill, Song-Chun Zhu, and Ying~Nian Wu.
\newblock Learning non-convergent non-persistent short-run mcmc toward
  energy-based model.
\newblock In {\em Advances in Neural Information Processing Systems}, pages
  5233--5243, 2019.

\bibitem{wavenet}
Aaron van~den Oord, Sander Dieleman, Heiga Zen, Karen Simonyan, Oriol Vinyals,
  Alex Graves, Nal Kalchbrenner, Andrew Senior, and Koray Kavukcuoglu.
\newblock Wavenet: A generative model for raw audio.
\newblock {\em arXiv preprint arXiv:1609.03499}, 2016.

\bibitem{radford2015unsupervised}
Alec Radford, Luke Metz, and Soumith Chintala.
\newblock Unsupervised representation learning with deep convolutional
  generative adversarial networks.
\newblock {\em arXiv preprint arXiv:1511.06434}, 2015.

\bibitem{rezende}
Danilo~Jimenez Rezende, Shakir Mohamed, and Daan Wierstra.
\newblock Stochastic backpropagation and approximate inference in deep
  generative models.
\newblock {\em arXiv preprint arXiv:1401.4082}, 2014.

\bibitem{salimans2016improved}
Tim Salimans, Ian Goodfellow, Wojciech Zaremba, Vicki Cheung, Alec Radford, and
  Xi~Chen.
\newblock Improved techniques for training gans.
\newblock In {\em Advances in neural information processing systems}, pages
  2234--2242, 2016.

\bibitem{pixelcnn++}
Tim Salimans, Andrej Karpathy, Xi~Chen, and Diederik~P Kingma.
\newblock Pixelcnn++: Improving the pixelcnn with discretized logistic mixture
  likelihood and other modifications.
\newblock {\em arXiv preprint arXiv:1701.05517}, 2017.

\bibitem{tanaka2019discriminator}
Akinori Tanaka.
\newblock Discriminator optimal transport.
\newblock In {\em Advances in Neural Information Processing Systems}, pages
  6813--6823, 2019.

\bibitem{turner2018metropolis}
Ryan Turner, Jane Hung, Eric Frank, Yunus Saatci, and Jason Yosinski.
\newblock Metropolis-hastings generative adversarial networks.
\newblock {\em arXiv preprint arXiv:1811.11357}, 2018.

\bibitem{pixelcnn}
Aaron Van~den Oord, Nal Kalchbrenner, Lasse Espeholt, Oriol Vinyals, Alex
  Graves, et~al.
\newblock Conditional image generation with pixelcnn decoders.
\newblock In {\em Advances in neural information processing systems}, pages
  4790--4798, 2016.

\bibitem{van2017neural}
Aaron van~den Oord, Oriol Vinyals, et~al.
\newblock Neural discrete representation learning.
\newblock In {\em Advances in Neural Information Processing Systems}, pages
  6306--6315, 2017.

\bibitem{10.5555/3104482.3104568}
Max Welling and Yee~Whye Teh.
\newblock Bayesian learning via stochastic gradient langevin dynamics.
\newblock In {\em Proceedings of the 28th International Conference on
  International Conference on Machine Learning}, ICML’11, page 681–688,
  Madison, WI, USA, 2011. Omnipress.

\bibitem{xiao2019generative}
Zhisheng Xiao, Qing Yan, Yi'an Chen, and Yali Amit.
\newblock Generative latent flow: A framework for non-adversarial image
  generation.
\newblock {\em arXiv preprint arXiv:1905.10485}, 2019.

\end{thebibliography}

\newpage
\appendix
\section{Proof of (\ref{reparam})}
\label{proof1}
Since $p_{\phi^*}$ is generated through a deterministic mapping $\phi^*$ from the
latent space $\mathcal Z$ to the observation space $\mathcal X$,
for any function $f$ on  ${\mathcal X}$ we have:
\begin{align*}
\mathbb{E}_{p_{\phi^*}}[f(\mathbf{x}] =& \int_{\mathcal X} f(\mathbf{x}) p_{\phi^*}(\mathbf{x}) d\mathbf{x}\\ = & \int_{\mathcal Z} f(\phi^*(\mathbf{z})) p(\mathbf{z}) d\mathbf{z} = \mathbb{E}_{p(\mathbf{z})}[f(\phi^*(\mathbf{z})].
\end{align*}
Therefore 
\begin{align*}
Z_{\phi^*,\theta} = \mathbb{E}_{p_{\phi^*}(\mathbf{x})}\left[e^{{-E_\theta(\mathbf{x})}}\right] = \mathbb{E}_{p(\mathbf{z})}\left[{e^{-E_\theta(G_{\phi^*}(\mathbf{z}))}}\right]
\end{align*}
Take derivative w.r.t $\theta$ we can get
\begin{align*}
    \frac{\partial\log Z_{\phi^*,\theta}}{\partial \theta} &= -\mathbb{E}_{p_{\phi^*,\theta}\left(\mathbf{x}\right)}\left[\frac{\partial E_{\theta}\left(\mathbf{x}\right)}{\partial \theta}\right]\\
    &= -\mathbb{E}_{p(\mathbf{z})}\left[\frac{e^{-E_\theta(G_{\phi^*}(\mathbf{z}))}}{Z_{\phi^*,\theta}}\frac{\partial E_\theta(G_{\phi^*}(\mathbf{z}))}{\partial \theta}\right]\\
    &= -\mathbb{E}_{p_{\phi^*,\theta}\left(\mathbf{z}\right)}\left[\frac{\partial E_{\theta}\left(G_{\phi^*}(\mathbf{z})\right)}{\partial \theta}\right],
\end{align*}
which is exactly (\ref{reparam}).

\section{Experimental Settings}\label{setting}
\subsection{Base generative models}
We first introduce training settings of the base generative models that we used in our experiments. We train GLOW basically following the settings provided in \cite{nalisnick2018deep}. For MNIST and Fashion MNIST,we use a GLOW architecture of 2 blocks of 16 affine coupling layers,
squeezing the spatial dimension in between the 2 blocks. For the coupling function, we use a 3-layer Highway network with 64 hidden channels. For CIFAR-10, we use 3
blocks of 32 affine coupling blocks, applying the multi-scale architecture between each block. The coupling function is a 3-layer Highway network with 256 hidden channels. Note that we modify the model size to fit in a single GPU for training. For MNIST and Fashion MNIST, we train the GLOW for 128 epochs with batch size 64 and Adam optimizer with fixed learning rate $5\times 10^{-4}$. For CIFAR-10, we train the GLOW for 256 epochs with batch size 64 and Adam optimizer with fixed learning rate $5\times 10^{-4}$.

Our use the DCGAN \cite{radford2015unsupervised} structure on the decoders of our VAEs, and the encoders are designed to be symmetric to the decoder. We use latent dimension $100$ for all experiments. For MNIST and Fashion datasets, we use binary cross entropy as reconstruction loss, while for CIFAR-10, we use MSE loss. All VAEs are trained for 256 epochs with batch size 128 and Adam optimizer with fixed learning rate $1\times 10^{-3}$. 

For GLF adopt the same encoder-decoder structure as in our settings for training VAEs. We use latent dimension $64$ for all experiments. The normalizing flow for matching the latent distribution is a simple GLOW network with 4 affine coupling layers, each consists of one fully connected layer with 256 units. The AE and the flow are jointly trained for 256 epochs with batch size 128 and Adam optimizer with fixed learning rate $1\times 10^{-3}$. 

\subsection{Energy based models}
We used a simplified version of the network structure described in \cite{du2019implicit} to define our $E_{\theta}$. In particular, our method consists of 3 resnet blocks with 64 hidden channels and 3 resent blocks with 128 hidden channels, followed by Global Sum Pooling and a FC layer. We also find the network structure in \cite{nalisnick2018deep}, which has much less parameters, leads to only slightly worse performances. Therefore, their energy function can be used for parameter efficiency.  

Unlike \cite{du2019implicit, nijkamp2019anatomy}
where the Langevin dynamics is dominated by the gradient, we find our latent EBMs work well with balanced noise and gradient in \eqref{z ld}. For Langevin dynamics, we use $\epsilon = 0.01$ and run the chain for 60 steps. We find adding a small amount (with coefficient $0.1$) of energy regularization is helpful for avoiding over-fitting early in the training. After training, we find sampling latent variables with longer chain leads to better performances. We generate samples from $p_{\phi^*,\theta}(\mathbf{z})$ by running the chain for 100 steps. 

For EBMs on the pixel space, we find short-run non-persistent training as described in \cite{nijkamp2019anatomy} hard to converge on MNIST and Fashion MNIST, so we follow the setting in \cite{du2019implicit}, where they use persistent initialization for the Langevin dynamics. They maintain a sample replay buffer during the training, and samples from the buffer are used to initialize the chain. We follow the hyper-parameter settings in \cite{du2019implicit},and we train EBMs on MNIST and Fashion MNIST for $20$k steps, and on CIFAR-10 for $100$k steps. Note that we train less number of steps on CIFAR-10 than the open source implementation of \cite{du2019implicit}, because it takes prohibitively long time due to the hardware constraint. We find our samples qualitatively comparable to those of \cite{du2019implicit} (see Figure \ref{ebm large}), but we are unable to match their reported FID scores on CIFAR-10, possibly due to not training long enough. After training, new samples are generated from chains initialized from the replay buffer.  
\section{Additional Qualitative Results}\label{qualitative}
\begin{figure*}[ht]
    \centering
    \begin{subfigure}{.48\linewidth}
    \includegraphics[scale=0.5]{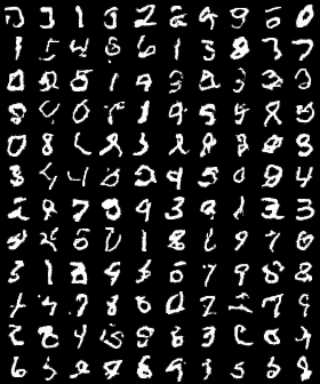}
    \end{subfigure}
    \begin{subfigure}{.48\linewidth}
    \includegraphics[scale=0.5]{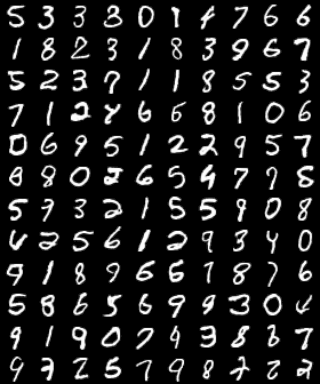}
    \end{subfigure}
     \vskip1em
     \begin{subfigure}{.48\linewidth}
    \includegraphics[scale=0.5]{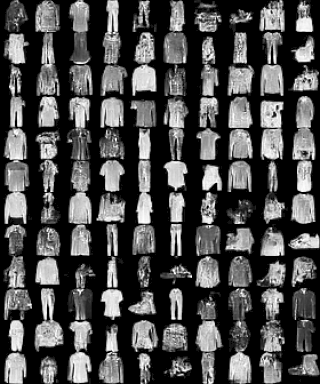}
    \end{subfigure}
    \begin{subfigure}{.48\linewidth}
    \includegraphics[scale=0.5]{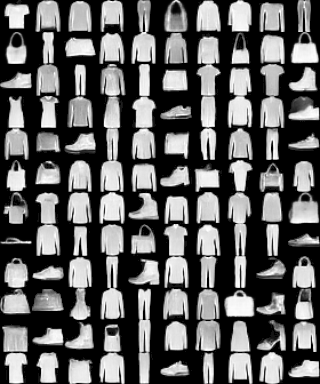}
    \end{subfigure}
    \vskip1em
     \begin{subfigure}{.48\linewidth}
    \includegraphics[scale=0.5]{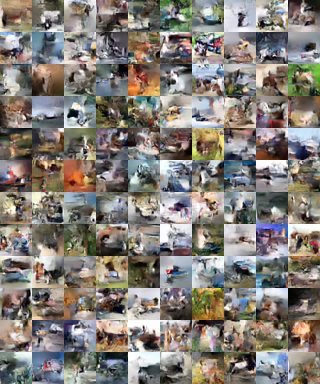}
    \end{subfigure}
    \begin{subfigure}{.48\linewidth}
    \includegraphics[scale=0.5]{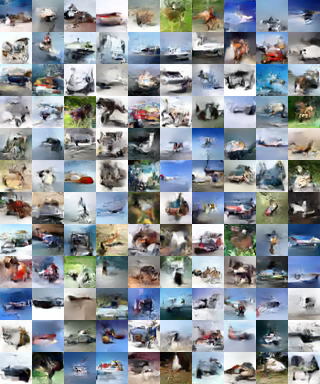}
    \end{subfigure}
    \caption{\label{flow large}
     Additional qualitative results of GLOW + atent EBM on MNIST, Fashion and CIFAR-10. \textbf{Left}: samples generated by $\mathbf{z}$'s from the prior. \textbf{Right}: samples generated by $\mathbf{z}$'s from $ p_{\phi^*,\theta}(\mathbf{z})$.}
\end{figure*}

\begin{figure*}[ht]
    \centering
    \begin{subfigure}{.48\linewidth}
    \includegraphics[scale=0.5]{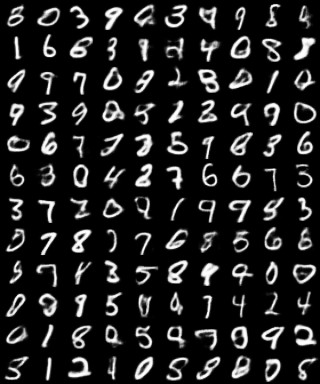}
    \end{subfigure}
    \begin{subfigure}{.48\linewidth}
    \includegraphics[scale=0.5]{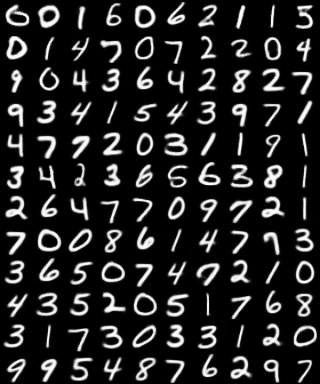}
    \end{subfigure}
     \vskip1em
     \begin{subfigure}{.48\linewidth}
    \includegraphics[scale=0.5]{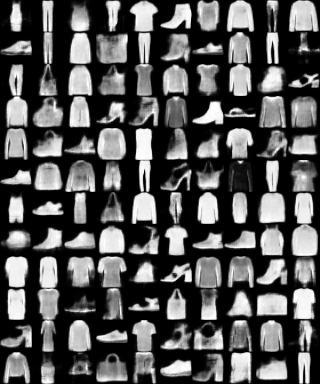}
    \end{subfigure}
    \begin{subfigure}{.48\linewidth}
    \includegraphics[scale=0.5]{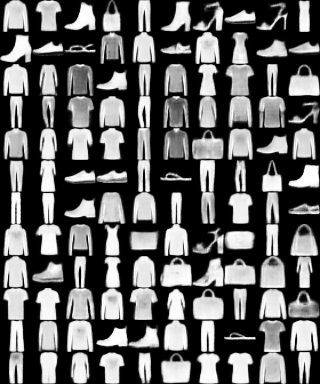}
    \end{subfigure}
    \vskip1em
     \begin{subfigure}{.48\linewidth}
    \includegraphics[scale=0.5]{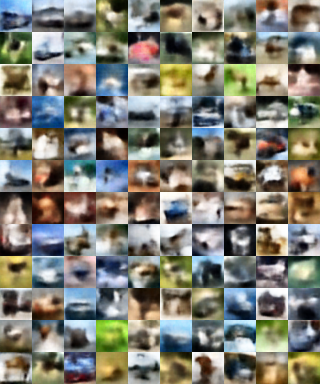}
    \end{subfigure}
    \begin{subfigure}{.48\linewidth}
    \includegraphics[scale=0.5]{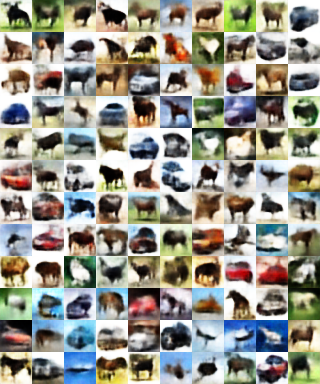}
    \end{subfigure}
    \caption{\label{vae large}
     Qualitative results of VAE + latent EBM on MNIST, Fashion and CIFAR-10. \textbf{Left}: samples generated by $\mathbf{z}$'s from the prior. \textbf{Right}: samples generated by $\mathbf{z}$'s from $ p_{\phi^*,\theta}(\mathbf{z})$.}
\end{figure*}

\begin{figure*}[ht]
    \centering
    \begin{subfigure}{.48\linewidth}
    \includegraphics[scale=0.5]{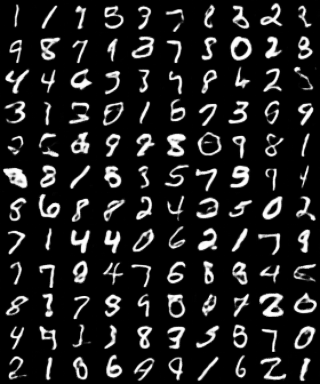}
    \end{subfigure}
    \begin{subfigure}{.48\linewidth}
    \includegraphics[scale=0.5]{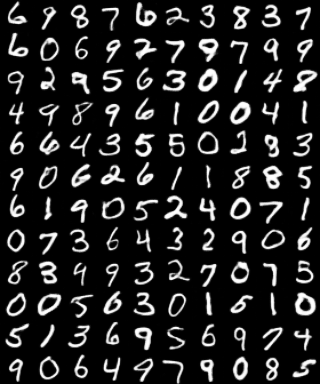}
    \end{subfigure}
     \vskip1em
     \begin{subfigure}{.48\linewidth}
    \includegraphics[scale=0.5]{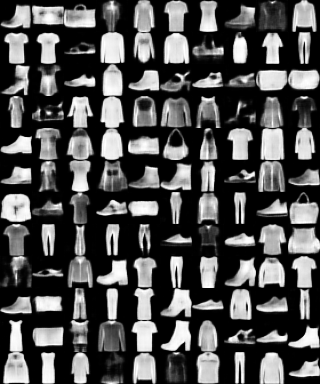}
    \end{subfigure}
    \begin{subfigure}{.48\linewidth}
    \includegraphics[scale=0.5]{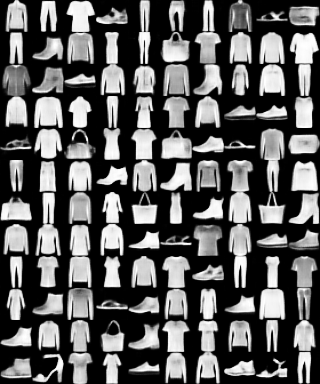}
    \end{subfigure}
    \vskip1em
     \begin{subfigure}{.48\linewidth}
    \includegraphics[scale=0.5]{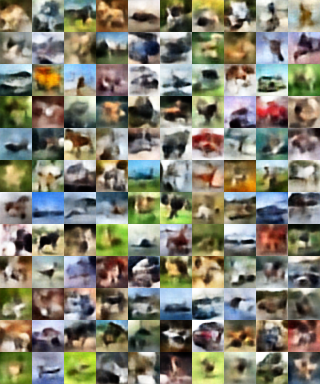}
    \end{subfigure}
    \begin{subfigure}{.48\linewidth}
    \includegraphics[scale=0.5]{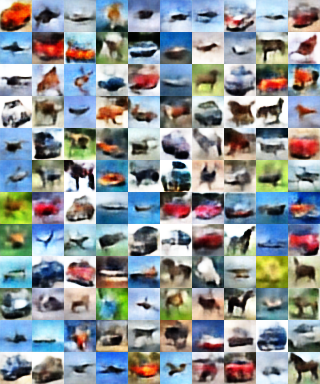}
    \end{subfigure}
    \caption{\label{glf large}
     Qualitative results of GLF + latent EBM on MNIST, Fashion and CIFAR-10. \textbf{Left}: samples generated by $\mathbf{z}$'s from the prior. \textbf{Right}: samples generated by $\mathbf{z}$'s from $ p_{\phi^*,\theta}(\mathbf{z})$.}
\end{figure*}

\begin{figure*}[ht]
    \centering
    \begin{subfigure}{.32\linewidth}
    \includegraphics[scale=0.39]{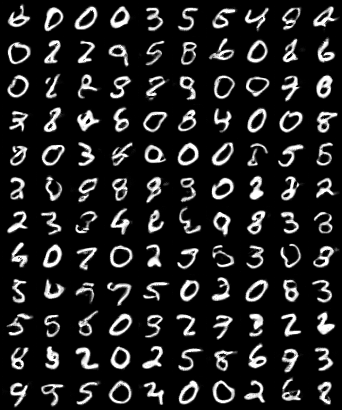}
    \caption{MNIST}
    \end{subfigure}
    \begin{subfigure}{.32\linewidth}
    \includegraphics[scale=0.39]{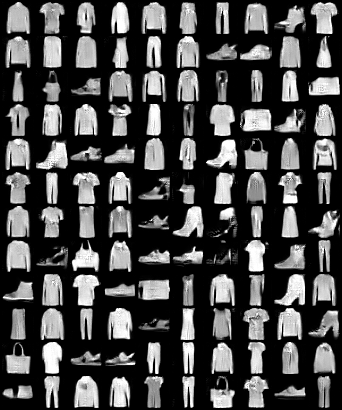}
    \caption{Fashion MNIST}
    \end{subfigure}
     \begin{subfigure}{.32\linewidth}
    \includegraphics[scale=0.41]{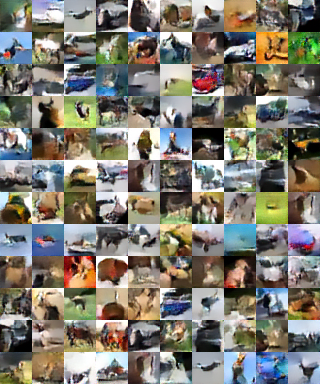}
    \caption{CIFAR-10}
    \end{subfigure}
    \caption{\label{ebm large}
     Qualitative samples from EBMs trained on pixel space.}
\end{figure*}

In this section, we show some additional qualitative results. In Figure \ref{flow large}, we presents more examples of samples from GLOW and GLOW + latent EBM, in addition to Figure \ref{glow fig} in the main text. In Figure \ref{vae large} we show samples from VAE and VAE + latent EBM. In Figure \ref{glf large}, we show samples from GLF and GLF + latent EBM. In all of these experiments, we clearly observe that latent EBMs improve the sample quality of base generative models. Finally, in Figure 
\ref{ebm large}, we show samples from EBMs trained on pixel space. 

\end{document}